# Adaptive Frequency Green Light Optimal Speed Advisory based on Hybrid Actor-Critic Reinforcement Learning

Ming Xu, *Member, IEEE*, Dongyu Zuo

*Abstract*—Green Light Optimal Speed Advisory (GLOSA) system suggests speeds to vehicles to assist them in passing through intersections during green intervals, thus reducing traffic congestion and fuel consumption by minimizing the number of stops and idle times at intersections. However, previous research has focused on optimizing the GLOSA algorithm, neglecting the frequency of speed advisory by the GLOSA system. Specifically, some studies provide speed advisory profile at each decision step, resulting in redundant advisory, while others calculate the optimal speed for the vehicle only once, which cannot adapt to dynamic traffic. In this paper, we propose an Adaptive Frequency GLOSA (AF-GLOSA) model based on Hybrid Proximal Policy Optimization (H-PPO) method, which employs an actor-critic architecture with a hybrid actor network. The hybrid actor network consists of a discrete actor that outputs control gap and a continuous actor that outputs acceleration profiles. Additionally, we design a novel reward function that considers both travel efficiency and fuel consumption. The AF-GLOSA model is evaluated in comparison to traditional GLOSA and learning-based GLOSA methods in a three-lane intersection with a traffic signal in SUMO. The results demonstrate that the AF-GLOSA model performs best in reducing average stop times, fuel consumption and CO2 emissions.

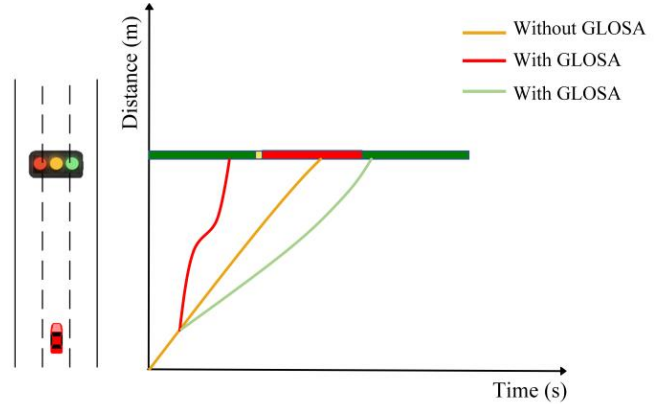

**Fig. 1.** Time-space diagram of intelligent vehicle.

## I. INTRODUCTION

THE combustion of fossil fuels in transportation has played a significant role in exacerbating global warming [1]. Consequently, there is a growing interest in developing intelligent transportation systems (ITS) and dynamic eco-driving to accelerate the deployment of low-carbon fuels [2]. Among the existing researches, Green Light Optimal Speed Advisory (GLOSA) is an efficient ITS service that provides optimal speed advisory profile to the intelligent vehicles to help them get through the intersection during the green phase. As shown in Fig 1, it assumes a traffic scenario that the intelligent vehicle reaches the intersection during red light (yellow line). If the intelligent vehicle could access real-time information about the driving condition of surrounding vehicles as well as upcoming traffic light timings in advance, then it can accelerate (red line) or decelerate (green line) in advance to cross the intersection without stopping.

Thanks to recent developments in ITS, the Connected and Autonomous Vehicles (CAVs) can obtain real-time traffic information through advanced communication technologies such as Vehicle to Vehicle (V2V) and Vehicle to Infrastructure (V2I). The CAVs equipped with GLOSA systems, utilizing advanced traffic information, can effectively reduce the occurrence of stop-and-go phenomena at intersections. This reduction leads to significant decreases in the number of stops [3], idle time [4], fuel consumption [5], and pollutant emissions [6] of intelligent vehicles.

In the U.S., traffic signal information has been provided to vehicles since 2016, indicating that such information will become increasingly accessible and widespread in the future [7]. And it also has been proven that providing real-time traffic information to vehicles can effectively alleviate traffic congestion [8] and reduce CO2 emissions [9].

In recent years, a large number of scholars have devoted themselves to the study of intelligent vehicle speed guidance, and we divide them into three approaches: rule-based methods, optimization-based method, and learning-based methods. Among that, the rule-based approach uses mathematical formulas and constraints to train strategies, which is easy to implement but this method needs expert knowledge and usually leads to local optimum, and this approach also cannot adapt to dynamic traffic. In order to obtain better performance, optimization-based methods are proposed to solve the defects

This work was supported in part by the Doctoral Scientific Research Foundation of Liaoning Technical University.

The Associate Editor for this paper war _ (Corresponding author: Ming Xu).

M. Xu is with the software college, Liaoning Technical University. (e-mail: xum.2016@tsinghua.org.cn).

D. Zuo is with the software college, Liaoning Technical University. (e-mail: 472121731@stu.lntu.edu.cn).

The source code is available at github: https://github.com/dongyu768/AF-GLOSA.

mentioned above. However, the defect of the optimization-based method is it requires a large amount of computing time consumption and fails to consider a variety of behaviors of the vehicles, such as longitudinal speed control, lateral lane change decision, or overtaking et. In order to meet the requirements of more real-time and dynamic performance, a learning-based approach is proposed to solve dynamic driving control studies.

For example, Yuan et al. [10] used the Deep Q-Network (DQN) method to control longitudinal speed of vehicles with the goal of reducing delays and travel time caused by stop-and-go behaviors at intersections. However, in this study, a rule-based approach is needed to ensure that the speed is changed only once throughout the entire decision step. Xia et al. [11] employed a parameterized reinforcement learning algorithm to train the agent for both longitudinal car-following control and lateral lane-changing decision. Similarly, Bai et al. [12] proposed a hybrid RL framework to generate both lateral and longitudinal eco-driving actions. Peng et al. [13] developed a double-layer decision-making model for controlling car-following and lane-changing operations using the D3QN and DDPG algorithms, respectively. Instead of offering a single advisory profile for the entire decision batch, [11], [12] and [13] all employ a purely learning-based approach and provide advisory profiles for vehicles at each decision step.

Based on the different types of advisory frequency above mentioned, we categorize speed advisory into three distinct patterns, as shown in Fig. 2. The GLOSA system provides vehicles with a single speed advisory and relies on a rule-based method to determine the optimal speed change mode (including acceleration, deceleration, or maintaining), as illustrated in Fig. 2(a). Upon entering the speed guidance area, the vehicle transmits its observation information (as detailed in Section III B) to the GLOSA system for analysis. The system evaluates the state information and determines whether the vehicle can pass through the intersection while maintaining its current speed during green light. If so, no speed advisory profile is provided. Otherwise, the system offers an optimal speed advisory profile to vehicles. However, this approach is insufficient in handling the dynamic and complex nature of traffic flow.

As depicted in Fig. 2(b), the system offers speed advisory to vehicles at each decision step. While this approach is adaptable to complex and dynamic traffic flow, it has a tendency to generate excessive speed advisory profiles. The high frequency of speed guidance may lead to increased computational overhead on systems utilizing traditional GLOSA algorithms, or result in higher fuel consumption for vehicles using learning-based GLOSA algorithms during the trip of vehicles. Moreover, frequent speed guidance may also reduce passenger comfort.

In this paper, we propose an Adaptive Frequency GLOSA (AF-GLOSA) model to address the limitations of the two speed advisory modes discussed above. As illustrated in Fig. 2(c), when a vehicle enters the speed guidance area, it transmits its current state to the GLOSA system, which evaluates whether the speed advisory is required based on the vehicle's current state. If speed advisory is necessary, the system provides an optimal speed advisory profile to the vehicle. Otherwise, the vehicle enters the speed advisory gap period and maintains its current driving state.

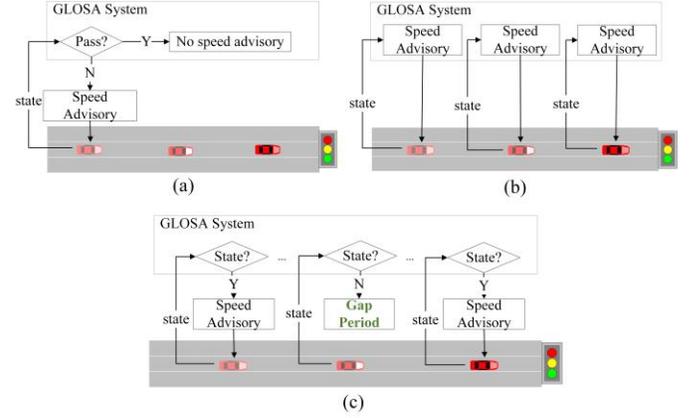

**Fig 2.** (a) Speed advisory at first decision step with rule-based method; (b) Speed advisory at each decision step; (c) Speed advisory with adaptive frequency.

In this paper, we formulate the frequency of speed advisory as a series of adaptive decisions and learn the relevant features from observed state information using a learning-based approach. This enables vehicles to dynamically adjust their driving state. Compared to the first two speed advisory modes, the third mode demonstrates superior adaptability to the dynamic changes in traffic flow, and when the traffic scene changes, the vehicle can promptly adjust its driving state in response. Finally, the AF-GLOSA model utilizes a hybrid actor-critic architecture to achieve both speed advisory frequency decision and longitudinal speed control for CAVs.

The main contributions of this paper can be summarized as follows:
- An Adaptive Frequency GLOSA (AF-GLOSA) model is proposed, which can dynamically adjust the speed advisory frequency based on dynamic traffic flow. To the best of our knowledge, we are the first to consider the impact of speed advisory frequency on GLOSA.
- We utilize the Hybrid Proximal Policy Optimization (H-PPO) method to implement AF-GLOSA model, which can effectively handle the parameterized action space task. Furthermore, we design a novel reward function that considers both travel efficiency and fuel consumption of vehicles.
- Finally, we evaluate the effectiveness of the proposed method in a three-lane intersection simulated scenario created by SUMO, with three different traffic densities.

II. RELATED WORK

Numerous studied have devoted themselves to improve the potentials of GLOSA system in increasing vehicles travel efficiency and saving energy. We have classified the existing researches into three categories based on different emphases: **(1) Developing and evaluating GLOSA system in the different traffic contexts**, including comparative evaluations of GLOSA effectiveness under single-vehicle and multi-vehicle scenarios, as well as evaluations of effectiveness under single-segment and multi-segment intersections. For example, Wu et al. [14] proposed a method for speed guidance in both single-vehicle and multi-vehicle scenarios, aiming to reduce delay and number of stops at the intersections. Seredynski et al. [15] compared single-segment and multi-segment GLOSA approaches and found that the multi-segment approach was more efficient in terms of travel time and fuel consumption. Similarly, Sharara et al. [16] also demonstrated that multi-segment approach outperforms single-segment approach. In addition, this paper discussed the influence of different activation distances on GLOSA. **(2) Algorithm optimization for different objectives.** In GLOSA research, algorithm optimization is usually performed to achieve specific targets, such as minimizing the number of stops, reducing travel time, decreasing fuel consumption, mitigating time delay, and minimizing pollutant emissions. For instance, in [17], an adaptive control algorithm is employed to optimize vehicle time loss, while [18] applied genetic algorithms to minimize both travel time and fuel consumption. [19] proposed a cooperative speed guidance algorithm that acquired traffic signal information at the approaching intersection to minimize delay and stop time. Furthermore, [20] reduced the travel time of vehicles by establishing GO and NO-GO indicators. **(3) Analysis the influence of external factors on GLOSA system performance.** These factors encompass the penetration rate of GLOSA-equipped vehicles, traffic density, the guidance area length, the speed guidance only to the lead vehicle in a platoon, the interaction between drivers' bounded rationality and speed guidance effectiveness, and others. For example, Katsaros et al. [21] proposed a target speed calculation method, and evaluated the average stop time and fuel consumption for GLOSA and non-GLOSA scenarios with different penetration rates of GLOSA-equipped vehicles and traffic density. The simulation results demonstrated that the benefits for fuel efficiency were more significant at higher penetration rates, while the benefits decreased with decreasing density. Ronald et al. [22] explored the optimal deceleration for reaching the stop line in both off-peak and peak-hour scenarios, and the results shown that the number of stop times increased during peak-hour scenarios. Stebbins et al. [23] proposed platoon-based GLOSA optimization algorithm, which sent suggested acceleration only to the lead vehicle of the platoon at each decision step, and the results showed that the algorithm could save the average fuel consumption between 21% and 23%. Tang et al. [24] demonstrated that driver's bounded rationality can impact fuel consumption and pollutant emissions during speed guidance. Feng et al. [25] studied the influence of platooning on GLOSA in order to increase intersection throughput.

The studies mentioned above have undoubtedly contributed to improving the efficiency of the GLOSA system in term of time and fuel consumption. However, the impact of speed advisory frequency on the GLOSA system has not been extensively investigated. In the previous studies, assuming the speed guidance time step is one second, the system provided speed advice in each second or only in the first second. Nonetheless, excessive frequency of speed guidance may increase the computational overhead of the system or result in higher fuel consumption during a trip, while insufficient frequency of speed guidance may decrease travel efficiency of vehicles or fail to adapt to dynamic traffic. Therefore, further research is needed to explore the influence of speed advisory frequency on the GLOSA system.

## III. Methodology

### A. Deep Reinforcement learning

Deep reinforcement learning (DRL) has been successfully applied in various domains such as robotics, games, and simulation environments. Reinforcement learning (RL) can be described as a Markov decision process (MDP) consisting of a five-tuple $\{S, A, P, \gamma, R\}$, where $S$ is the set of states, $A$ is the set of actions, $P$ is the set of transition possibility from state $s_i$ to state $s_{i+1}$, $\gamma$ is the discount factor, $\gamma \in [0, 1]$, R is the set of rewards. In general, the policy $\pi$ is a mapping from states to a probability distribution over actions: $\pi: S \to p(A = a|S)$. The objective of RL is to learn an optimal policy $\pi$ that allows an agent to select the best action $a$ based on the current state $s_i$, informed by the reward signals $r$ obtained from the environment.

The policy learning methods can be divided into three classes: value-based methods, policy-based methods and hybrid methods that use both. Deep Q-Network (DQN) is a widely used value-based method in DRL that leverages deep neural networks to model value functions and can handle high-dimensional and nonlinear state spaces while exhibiting good generalization capabilities. Proximal Policy Optimization (PPO) [26] is a reinforcement learning algorithm that simultaneously learns both policy and value functions based on the actor-critic framework. In contrast to other actor-critic methods, such as trust region policy optimization (TRPO) method, PPO utilizes a clipped surrogate objective to limit the difference between the new and old policies. The approach is simpler and more intuitive and demonstrates higher stability and performance in practical applications. In our method, we employ the Hybrid Proximal Policy Optimization (H-PPO) method, which is an extension of PPO designed to handle parameterized action space tasks.

The H-PPO [27] takes the hybrid actor-critic architecture and uses PPO as the policy optimization method for both its discrete policy $\pi_{\theta_d}$ and continuous policy $\pi_{\theta_c}$. The $\pi_{\theta_d}$ and $\pi_{\theta_c}$ are updated separately by minimizing their respective clipped surrogate objective. The objective for the discrete policy $\pi_{\theta_d}$ is the following:

$$L_d^{CLIP}(\theta_d) = \hat{E}_t[min\,(r_t^d(\theta_d)\hat{A}_t, clip(r_t^d(\theta), 1 - \epsilon, 1 + \epsilon)\hat{A}_t] \quad (1)$$

and the objective for the continuous policy $\pi_{\theta_c}$ is the following:

$$L_c^{CLIP}(\theta_c) = \hat{E}_t[min\,(r_t^c(\theta_c)\hat{A}_t, clip(r_t^c(\theta), 1 - \epsilon, 1 + \epsilon)\hat{A}_t] \quad (2)$$

where $\hat{A}_t$ is the advantage function:

$$\hat{A}_t = r_t + \gamma V(s_{t+1}) - V(s_t) \quad (3)$$

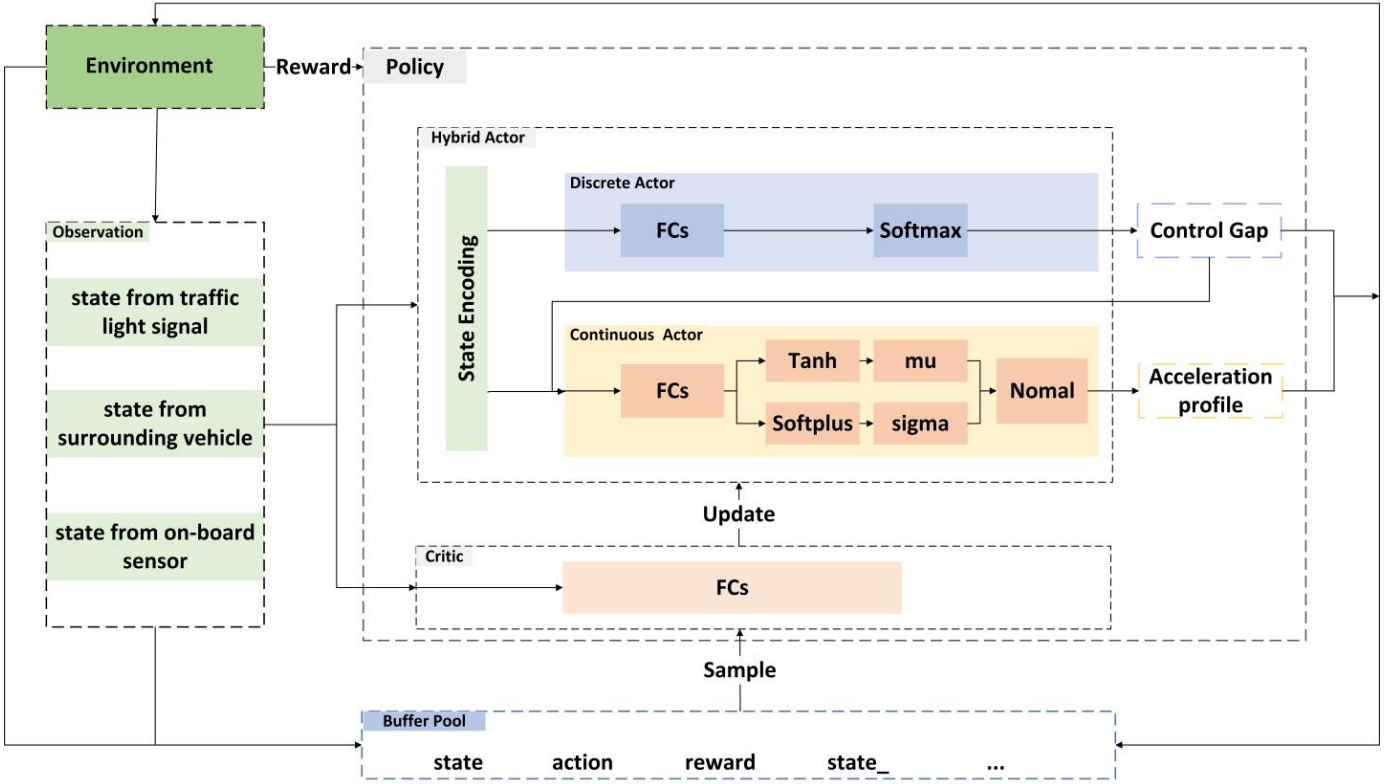

**Fig. 3.** The framework of AF-GLOSA model.

$\epsilon$ is a hyperparameter, and $\gamma$ is discount factor. Finally, $r_t^d(\theta_d)$ denotes the probability ratio of $\pi_{\theta_d}$, and is defined as

$$r_t^d(\theta_d) = \frac{\pi_{\theta_d}(a_t|s_t)}{\pi_{\theta_{d(old)}}(a_t|s_t)} \quad (4)$$

$r_t^c(\theta_c)$ denotes the probability ratio of $\pi_{\theta_c}$, and is defined as:

$$r_t^c(\theta_c) = \frac{\pi_{\theta_c}(a_t|s_t)}{\pi_{\theta_{c(old)}}(a_t|s_t)} \quad (5)$$

*B. Framework of Adaptive Frequency GLOSA*

In this section, we propose an Adaptive Frequency Green Light Optimized Speed Advisory (AF-GLOSA) model for CAVs, which employs hybrid proximal policy optimization (H-PPO) method to adapt to parameterized action space. As shown in Fig. 3, the intersection environment, similar to the scenario depicted in Fig. 4, treats the CAVs equipped with the GLOSA system as agent that can obtain observation information from surrounding environment, including incoming traffic light phase information, driving state information of surrounding vehicles, and speed and position information from onboard sensors. The AF-GLOSA model processes the observation information and outputs the control gap and acceleration advisory profile. The agent then performs the advisory actions to obtain the next state, and the environment scores the actions based on the pre-designed reward function.

The buffer pool is utilized to store trajectory information, which includes the initial observation of the agent, the speed advisory action generated by the AF-GLOSA model, the reward value received from the environment, and the subsequent observation of the agent after executing the advisory actions. The buffer pool serves as a data buffer for learning, and once it reaches its capacity, the trajectory information is utilized to train the policy $\pi_{\theta_d}$ and $\pi_{\theta_c}$. Subsequently, the buffer is cleared until it reaches its capacity again for the next training cycle.

To training the agent learns an optimal strategy, the state space, action space and reward function is designed as follows.

**State space.** State space is defined as the set of all possible states that an agent can be in at a given time step. In this context, the agent needs to select the optimal action based on real-time state information acquired from the road network environment, including controlled vehicles, traffic lights, and intersections. The state space $S_t$ can be formulated as follow:

$$S_t = [l_t, v_t, a_t, m_t, w_t, pre_v, pre_d, p_t] \quad (6)$$

where $l_t$ is the distance between the CAV and the stop line ahead, while $v_t$ and $a_t$ represent the current speed and acceleration of the controlled vehicle at time step $t$, $p_t$ is the current phase of the traffic light, $m_t$ represents the remaining time of the current phase, and $w_t$ is the waiting time for the front light to turn green. Additionally, $pre_v$ and $pre_d$ represent the speed of the preceding vehicle and the distance between the CAV and the preceding vehicle. In this paper, we just consider the longitudinal speed guidance, so we also just consider the driving state of preceding vehicle. Take an example, the sequence $[l_t, v_t, a_t, m_t, w_t, pre_v, pre_d, p_t] = [30, 8, 2.5, 2, 22, 11, 8, 0]$ indicates that the distance between the CAV and stop line is $30\,m$, the current speed and acceleration of CAV are $8m/s$ and $2.5m/s^2$, respectively. The speed of the preceding vehicle is $11m/s$, the distance between CAV and the preceding vehicle is $8m$, the current phase of incoming traffic light is green, the remaining time of the current phase is $2s$, and the waiting time for the incoming light to turn green is $22s$.

**Parameterized action space.** We design a parameterized action space, consisting of two sub-action spaces, to accomplish the adaptive frequency green light optimized speed advisory task.

The first sub-action space is the discrete speed control gap action space, denoted as $A_{sc}$:
$$A_{sc} = \{0, 1\} \quad (7)$$

If the speed control gap action is set as 1, the AF-GLOSA model will proceed to the next action space for acceleration advice. Otherwise, the controlled vehicle will enter speed control gap period and maintain current driving state.

The second sub-action space is the continuous acceleration-control action space, denoted as $A_{ac}$:
$$A_{ac} = \{a \mid a \in [-3, 3]\} \quad (8)$$

The maximum acceleration and deceleration of the controlled vehicles are set as $3\ m/s^2$ each. The AF-GLOSA model is designed to generate an optimal acceleration profile that falls within this range.

**Reward function.** The reward function is a critical component in reinforcement learning, which provides the agent with necessary feedback signal to evaluate its actions and guide its decision-making towards achieving desired goals. It defines the objective of the task and significantly influences the agent's behavior and learning speed. Designing an appropriate reward function is a challenging and important task in reinforcement learning. In this paper, the optimization goal is to reduce the number of stops and fuel consumption of CAVs. Therefore, the reward function $R$ is designed as follow:
$$R = \alpha * r_1 + \beta * r_2 + \omega * r_3 \quad (9)$$

where $r_1$ represents the fuel consumption of controlled vehicle in a decision step, and $r_2$ is set based on the number of stops to reward or punish the agent, and the $r_2$ is formulated as follow:
$$r_2 = \begin{cases} -200, & \text{if } v_t \leq 0.1 \\ 0, & \text{else} \end{cases} \quad (10)$$

The weight coefficients $\alpha$, $\beta$ and $\omega$ are used to balance the different reward items. And the $r_3$ is a reward item aiming to encourage the agent to choose an ideal acceleration.

Algorithm 1 outlines the calculation process of the reward item $r_3$. We differentiate the acceleration advice into two categories: **appropriate** and **inappropriate**. An acceleration advisory that recommends accelerating when the vehicle is already traveling at the maximum speed is considered inappropriate, as well as an acceleration guidance that leads to an unreasonable speed beyond the preset upper and lower limits. In Algorithm 1, if the controlled vehicle is in the control period, we calculate the target speed $v_{aim}$ based on the advised acceleration. If the acceleration is appropriate (line:3-5), $r_3$ will be assigned a positive value as reward; otherwise, it will be assigned a negative value as punishment (line:6-7). Conversely, if the vehicle is within the control gap period (line:8-11), $r_3$ will be given a positive reward. However, if the

---

**Algorithm 1** $r_3$ reward item setting

**Input**: ego-vehicle's speed $v_t$, control gap $a_{dis}$
**Output**: $r_3$ reward value
1: **If** $a_{dis}$ **do**
2:     calculate target speed $v_{aim}$ based on acceleration advisory $a_{adv}$
3:     **If** $v_{aim}$ between $v_{min}$ and $v_{max}$ **do**
4:         set $r_3$ as 5
5:         set the ego-vehicle's target speed as $v_{aim}$
6:     **else**
7:         set $r_3$ as -2
8: **else**
9:     set $r_3$ as 4
10:     **If** $v_t$ is less than $v_{min}$
11:         set $r_3$ as -2
12: **Return** $r_3$

---

current speed is deemed inappropriate (line:10-11), indicating the need for a speed advice in this decision step, $r_3$ will be penalized with a negative value.

The reinforcement learning part of the AF-GLOSA model is composed of a hybrid actor network and a critic network. The hybrid actor consists of a discrete actor and a continuous actor, both of which enjoy an Encode layer. The discrete actor decides whether the agent should enter speed control gap period, aiming to give the optimal advisory frequency in the whole decision step, while the continuous actor provides the optimal acceleration profiles. As shown in Fig. 3, the observation is first inputted into the Encode layer, followed by fully connected layers, and finally is randomly sampled from the Softmax distribution to output control gap decisions. The continuous actor network takes the control gap decision and encoded state as inputs. Unlike the discrete actor, the continuous actor uses Normal distribution sampling and aims to output acceleration profiles. To limit the sampling range of the Normal distribution, we apply Tanh activation function to restrict the mean value of the Normal distribution between maximum acceleration and deceleration. The network architecture of the AF-GLOSA model is shown in Table I.

IV. SIMULATION ANALYSIS AND RESULT

*A. Scenario Description*

The study scenario is a three-lane urban intersection with a traffic light, as shown in Fig 4. The red vehicle represents the connected and automated vehicle (CAV) that is controlled by the GLOSA model, while the yellow vehicles represent human-driving vehicles (HDVs) that are controlled by the SUMO simulator. When the CAV enters the speed guidance area (the green section), it can receive information from the incoming traffic light and the preceding vehicle. Based on this real-time information, the CAV can adjust its speed to pass through the intersection without stopping.

In order to simulate the diversity of the driving vehicles,

TABLE I
DECISION MAKING ARCHITECTURE OF AF-GLOSA

| Function Layers | Input | Output |
|---|---|---|
| Encode layer | 8 | 128 |
| Discrete Actor | 128 | 2 |
| Continuous Actor | 128+1 | 1 |
| Critic Layer | 8 | 1 |

we have set up a total of five different types of HDVs with varying acceleration and deceleration capabilities [12], as shown in Table II. By setting different vehicle parameters, we can simulate the performance of HDVs under different traffic conditions more comprehensively. The vehicle length is fixed at 5 meters, while the minimize gap is set at 2 meters to maintain a stable preset headway and avoid collisions. The maximum speed of the vehicles is set at 11 m/s (40km/h). To simulate the behavior of HDVs, we adopt the Intelligent driver model (IDM) [28] follower model in SUMO. Finally, the probability parameter represents the likelihood of generating a simulated HDV.

It should be noted that the signal light timing in our simulation scenario adopts a traditional approach, where the green light phase duration $t_g$ is set to 20 seconds, and the red-light phase duration $t_r$ is also set to 20 seconds, result in a cycle time of the signal light $c$ of 40 seconds. For the sake of simplicity and to reduce training complexity, the yellow light phase has been excluded from our experimental protocol.

*B. Simulation setting*

Simulation of Urban Mobility (SUMO) [29] is a widely used open-source traffic simulator that facilitates modeling and simulation of urban traffic and vehicles' microscopic behaviors. The Traci interface within SUMO enables the retrieval and modification of simulation values using various programming languages. In this study, we build a signalized intersection scenario in the SUMO environment, similar to the scenario depicted in Fig. 4. We performed the simulation using SUMO version 1.15 and python version 3.7 on a system equipped with an AMD Ryzen 7 4800H with Radeon Graphics 2.90 GHz and an NVIDIA GeForce GTX 1650 GPU. The simulation parameters are listed in Table III, while the parameters for the CVAs are presented in Table IV.

*C. Simulation experiment and results*

To evaluate the performance of the AF-GLOSA model, we compared it with traditional and learning-based methods, including the following:

**Benchmark:** A test drive without using any GLOSA algorithms.

**S_GLOSA** [17]**:** The traditional GLOSA algorithm used by SUMO.

**L_GLOSA:** A learning-based GLOSA algorithm trained using the PPO method, which does not incorporate adaptive frequency.

AF-GLOSA: Learning-based Adaptive Frequency GLOSA

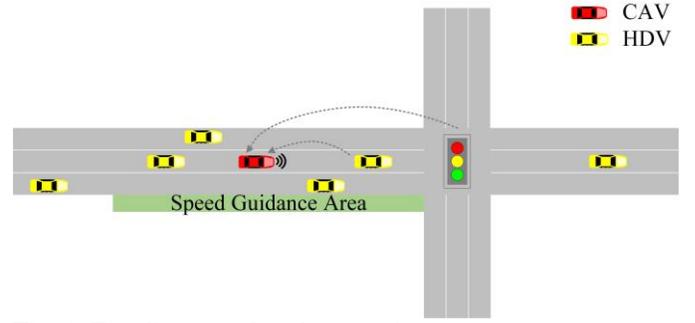

**Fig. 4.** Toy diagram of study scenario.

TABLE II
THE PARAMETER SETTING OF HDVS

| Vehicle | Acceleration | Deceleration | Length | Min Gap | Probability |
|---|---|---|---|---|---|
| **Type A** | 6.0 | 6.0 | 5 | 2 | 0.1 |
| **Type B** | 5.0 | 4.5 | 5 | 2 | 0.2 |
| **Type C** | 3.0 | 5.0 | 5 | 2 | 0.3 |
| **Type D** | 3.0 | 3.0 | 5 | 2 | 0.3 |
| **Type F** | 2.0 | 1.5 | 5 | 2 | 0.1 |

TABLE III
THE PARAMETER SETTING OF SIMULATIONS

| Item | Value | Unit |
|---|---|---|
| The number of lanes | 3 | - |
| The length of route | 994.9 | m |
| The duration time of green phase | 20 | s |
| The duration time of red phase | 20 | s |
| The road speed limit | 11 | m/s |
| The detected length of detector | 300 | m |
| Signal cycle | 40 | s |

TABLE IV
THE PARAMETER SETTING OF CAVS

| Item | Value | Unit |
|---|---|---|
| The minimized control speed $v_{min}$ | 4 (14.4) | m/s (km/h) |
| The maximized control speed $v_{max}$ | 11 (40) | m/s (km/h) |
| The length of speed guidance zone | 240 | m |
| The maximized acceleration $a_{max}$ | 3 | $m/s^2$ |
| The maximized deceleration $d_{max}$ | 3 | $m/s^2$ |

method that we proposed in this paper.

To compare the experimental performance of different algorithms, we defined four evaluation indicators as follows:

**WTI**: Waiting time (WTI) represents the total time that the vehicle has spent waiting during its trip.

When a vehicle is stopped at a location due to traffic congestion or traffic signal control, the WTI parameter increases by the length of time that the vehicle is stopped and waiting in seconds. Therefore, WTI can provide information on the extent to which a vehicle is affected by traffic congestion and traffic signal control during its trip.

TABLE V
EXPERIMENTAL PERFORMANCE OF DIFFERENT METHODS

| METHOD | WTI | | | WCO | | | CO2 | | | FUEL | | |
|---|---|---|---|---|---|---|---|---|---|---|---|---|
| | 300 | 1200 | 2700 | 300 | 1200 | 2700 | 300 | 1200 | 2700 | 300 | 1200 | 2700 |
| **Benchmark** | 7.58 | 5.75 | 6.11 | 0.8 | 0.8 | 0.71 | 254441 | 250378 | 254020 | 81157 | 79861 | 81023 |
| **S_GLOSA** | 7.18 | 5.5 | 5.94 | 0.73 | 0.73 | 0.71 | 251103 | 246776 | 251363 | 80092 | 78712 | 80176 |
| **imp.** | -0.4 | <u>-0.25</u> | -0.17 | -0.07 | -0.07 | -0 | <u>-3338</u> | <u>-3602</u> | <u>-2656</u> | <u>-1065</u> | <u>-1149</u> | <u>-847</u> |
| **L_GLOSA** | 5.91 | 6.09 | 4.53 | 0.69 | 0.88 | 0.62 | 267973 | 269633 | 263128 | 85473 | 86003 | 83928 |
| **imp.** | <u>-1.67</u> | +0.34 | <u>-1.58</u> | <u>-0.11</u> | +0.08 | <u>-0.09</u> | +13532 | +19255 | +9108 | +4316 | +6142 | +2905 |
| **AF-GLOSA** | 1.25 | 1.75 | 3.13 | 0.23 | 0.35 | 0.35 | 229354 | 227379 | 229734 | 73155 | 72525 | 73276 |
| **imp.** | **-6.33** | **-4.00** | **-2.98** | **-0.57** | **-0.45** | **-0.36** | **-25087** | **-22999** | **-24286** | **-8002** | **-7336** | **-7747** |

**WCO**: Waiting count (WCO) represents the number of times that the vehicle has stopped and waited during its trip. When a vehicle is stopped at a location due to traffic congestion or traffic signal control, the WCO parameter increases by 1.

**CO2**: The CO2 emissions model [30], in sumo, is based on the "COPERT IV" model, which is a widely used emissions model for road transport, which can be used to evaluate the environmental impact of different traffic scenarios and to assess the effectiveness of strategies aimed at reducing emissions, and the unit is milligram.

**FUEL**: The amount of fuel consumed by the vehicle in milligram.

To ensure the robustness and diversity of the simulations, we set traffic flow density at three levels: low (300 vehicles/hour), medium (1200 vehicles/hour), and high (2700 vehicles/hour). We also randomly selected the initial phase of traffic signal and departure time of the vehicles to avoid overfitting.

Finally, we tested the scalability of the AF-GLOSA model by varying the departure time range, and the experimental results were obtained by averaging over 100 random tests, as shown in Table V. To ensure fairness, we set the same random seed for each of the contrast methods.

For the case of 300 vehicles/hour in Table V, we present the results of the Benchmark method, which did not use any speed guide, for the CAV driving through the entire route. The waiting time of the vehicle was 7.58 s, and the WCO was 0.8, indicating that it stopped 80 times in all 100 tests. During the entire trip, a total of 254441 milligrams of CO2 were emitted, and the total fuel consumption was 81157 milligrams. In this context, we compared the performance of the other three methods. The improvement (imp.) rows show the relative improvement of each method over the Benchmark, while bold and underline highlight the best and second-best methods, respectively. Based on the experimental results, the traditional method was found to be capable of reducing the number of stops, fuel consumption, and emission to vehicles at a limited extent, with excellent stability. However, the effectiveness of this method decreased as the traffic flow increased. On the other hand, the learning-based method outperformed the traditional method in several indicators, but its performance was not stable enough. Furthermore, due to the lack of consideration of the frequency of vehicle advisory, the learning-based method led to an increase in fuel consumption and pollutant emission. Conversely, the proposed adaptive frequency GLOSA model demonstrated significant improvements across various traffic flows, exhibiting good stability and scalability.

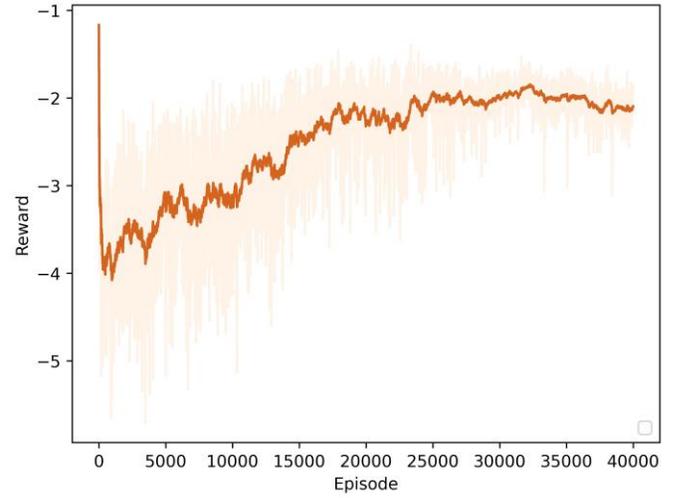

Fig. 5. The convergence process of reward value.

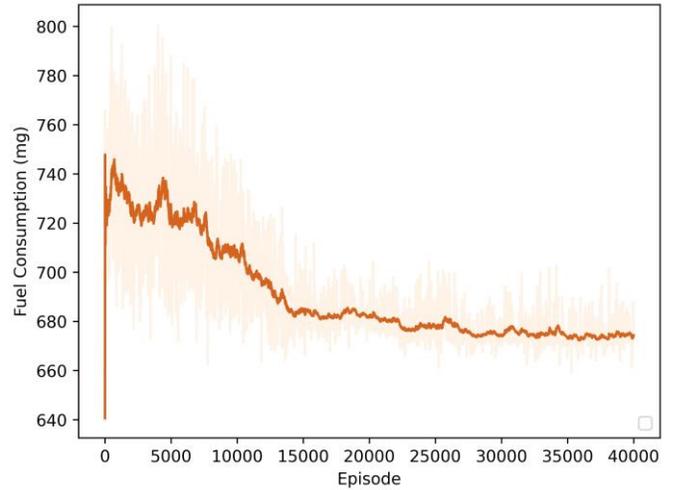

Fig. 6. The convergence process of fuel consumption.

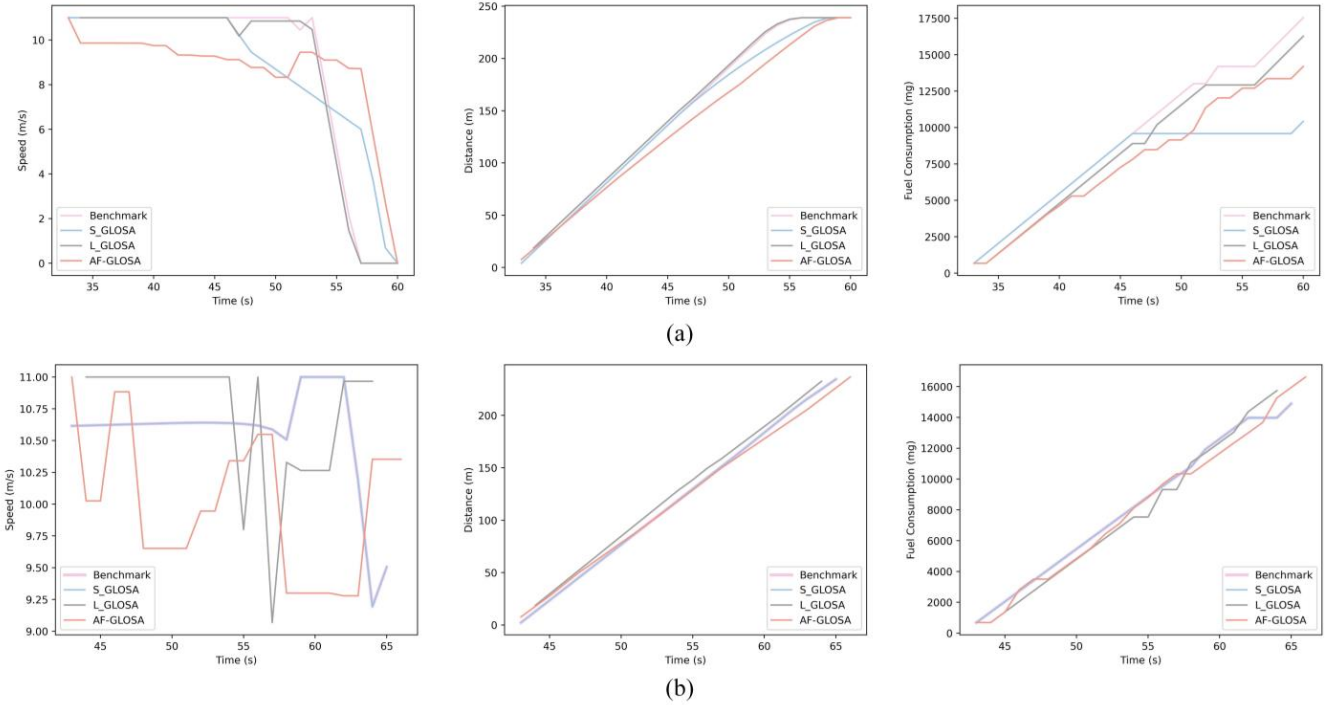

**Fig. 7.** Spatial-temporal diagrams of the CAV under different traffic scenarios; (a) Scenario 1: upcoming traffic signal is red light; (b) Scenario 2: upcoming traffic signal is green light.

TABLE VI
HYPER-PARAMETER SETTING OF OUR METHOD

| Parameter | value |
|---|---|
| Number of training episodes | 40000 |
| Learning rate for discrete actor | 3e-5 |
| Learning rate for continuous actor | 3e-5 |
| Learning rate for critic | 0.001 |
| Buffer size | 10000 |
| Batch size | 256 |
| Minimize batch size | 8 |
| Clipping $\epsilon$ | 0.1 |

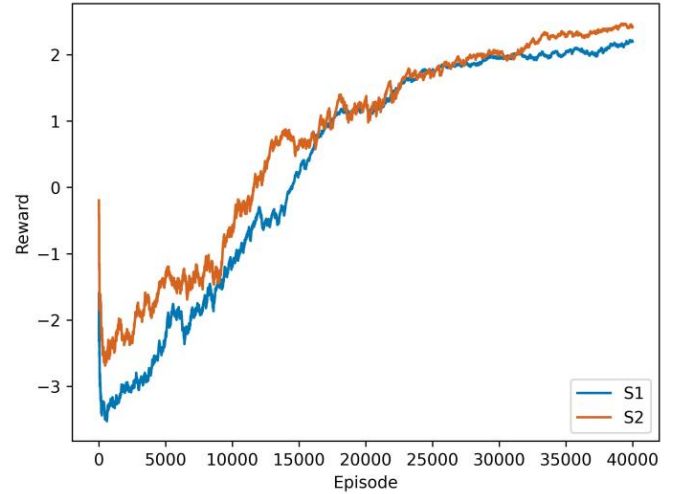

**Fig. 8.** The influence of different waiting time setting.

The training process involved 40,000 steps, and the convergence process of the reward function is depicted in Fig. 5. The fuel consumption displayed a downward trend and gradually converged during the entire training process, as illustrated in Fig. 6.

We also evaluated the performance of four methods in two traffic scenarios. The first scenario is that the vehicle will approach a red light without using the GLOSA system, as illustrated in Fig. 7 (a). The second scenario is that the vehicle can pass through the intersection during the green light phase without using the GLOSA system, as shown in Fig. 7 (b). We choose the different scenarios with choosing different depart time of CAVs.

In each traffic scenario, we respectively compared the four methods in terms of speed, distance from stop line and overall fuel consumption over time. According to the test results, our proposed method can better adjust the vehicle's speed and enable the vehicle passing through the intersection smoothly when speed advisory is needed (scenario 1). However, when the vehicle can cross the intersection without acceleration advisory (scenario 2), the traditional GLOSA method remains unresponsive, and the change trend of each indicator remains the same as that of the Benchmark. On the other hand, the learning-based method shows slight instability in the speed change.

Finally, the hyper-parameter setting of our method are shown in Table VI.

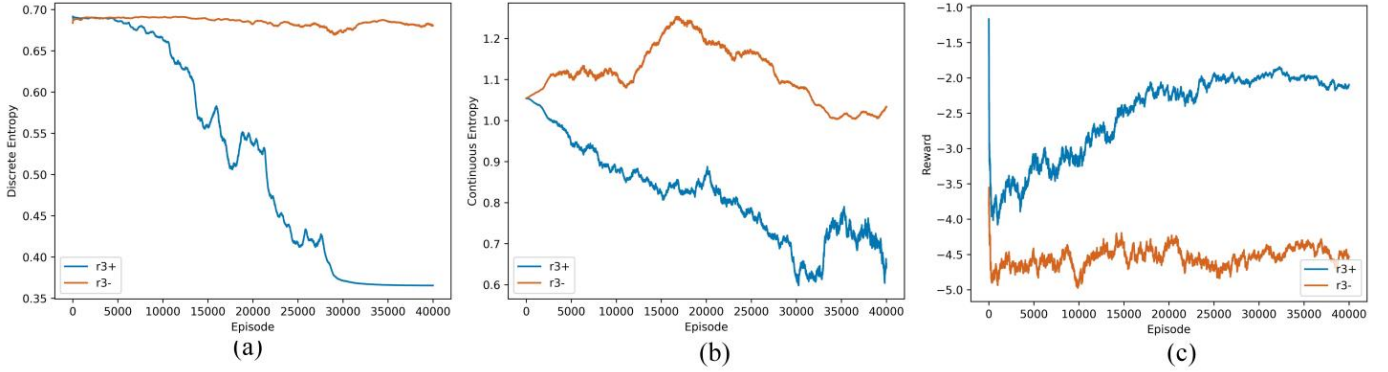

**Fig. 9.** The influence of adding $r_3$ item and removing $r_3$ item; (a) The entropy of discrete actor; (b) The entropy of continuous actor; (c) The reward value.

*D. Contrast experiment on state space design*

The design of state space can have a significant impact on experimental results. During our experiments, we found some ambiguity regarding the waiting time $w_t$ for the incoming traffic light to turn green. Assuming that the traffic light in front of the CAV is labeled as $L$, when the phase of $L$ is red, the value of $w_t$ is straightforward as it equals the remaining time of the red phase. However, when the current phase of $L$ is green, we encountered two possible situations for setting $w_t$. In Situation 1 (S1), we set $w_t = 0$, indicating that $w_t$ is the waiting time for $L$ to turn green. In Situation 2 (S2), we set $w_t = m_t + t_r$, which is the remaining time of the green phase plus the duration time of the red phase. In this case, $w_t$ represents the waiting time for $L$ to turn green again.

In order to compare the influence of different states on training, the experimental results of the two situations are shown in Fig. 8. And the other conditions are the same. As can be seen from the change of reward value, S2 is better than S1.

*E. Contrast experiment on reward function design*

The key aspect of designing the reward function is to balance the coefficient weight between fuel consumption and vehicle stopping counts, represented by $\alpha$ and $\beta$ respectively.

In this paper, we set $\alpha$ as -0.1, $\beta$ set as 0.6 and $\omega$ set as 10. Additionally, $r_3$ is a reward item that constrains the policy output edge values. Taking the continuous actor output as an example, the continuous actor network trends to output acceleration edge values that either maximize acceleration or maximize deceleration. We also conducted a comparison between adding and removing $r_3$ item respectively. As shown in Fig. 9, Fig. 9(a) and Fig. 9(b) represent the entropy of the output action for the discrete actor and continuous actor respectively. The level of entropy is an indicator of the stability of the network output action, with smaller entropy indicating better performance. The results show that adding the reward term $r_3$ leads to a reduction in entropy, and $r_3$ item also improves convergence in reward value, as demonstrated in Fig. 9(c).

*F. Contrast experiment on different simulation step*

The control step represents the time interval required for the speed advisory system to control the vehicle to the target speed, so an appropriate control step is necessary.

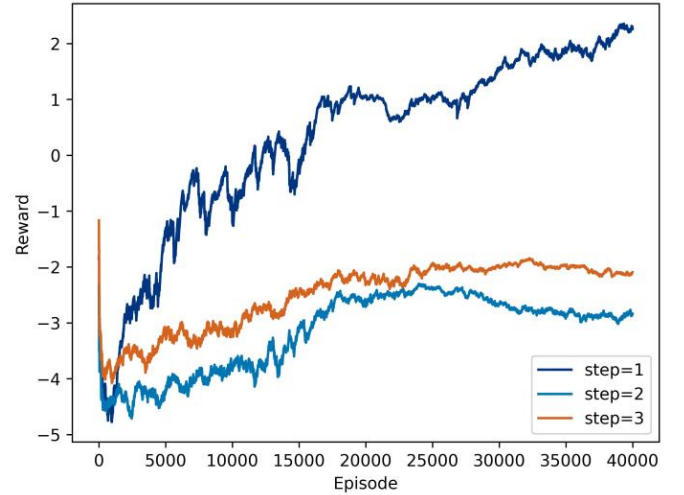

**Fig. 10.** The influence of control step.

TABLE VII
THE DIFFERENCE OF CONTROL STEP

| STEP | WTI | WCO | CO2 | FUEL |
|------|------|------|--------|-------|
| 1 | 1.81 | 0.27 | 235545 | 75130 |
| 2 | 1.25 | 0.23 | 229354 | 73155 |
| 3 | 1.38 | 0.21 | 214607 | 68451 |

In this study, we divided the control step into three levels, namely step=1, 2, 3. Figure. 10 shows the comparison of the reward function at different control steps over 40000 episodes. We observed that the reward function was highest at control step equals 1, but it results in significantly increased in fuel consumption and CO2 emissions compared to other control steps. To evaluate the performance of different control step models, we conducted experiments and presented the results in Table VII. We found that control step equals 3 achieved the best performance in three evaluation indicators.

V. CONCLUSION AND FUTURE WORK

In this paper, we demonstrated the importance of speed advisory frequency in GLOSA systems and proposed an Adaptive Frequency GLOSA model that utilizes deep reinforcement learning method to determine suitable advisory frequency and acceleration advisory profiles. The simulation experiments show that our model has achieved good performance in both traffic efficiency and energy consumption of intelligent vehicles and our method also has good scalability under different traffic densities.

Furthermore, the method that we proposed adaptive frequency approach can also be applied to other research areas in intelligent vehicle driving behavior decision-making tasks. In the future, our research will focus on optimizing travel efficiency in the context of multi-vehicle cooperation.

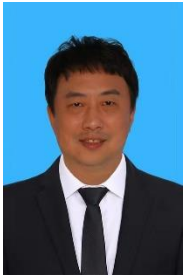

**Ming Xu** received his Ph.D. degree in computer science from Beijing University of Posts and Telecommunications, Beijing, China, in 2015. From 2016 to 2019, he was a Post-Doctoral Fellow with Tsinghua University, Beijing, China.

He is currently a professor with the software college, Liaoning Technical University. He has published over 20 papers in journals and conferences, including TITS\BIBM\TCSS. His research work has reported by MIT Technology Reviews. He is the recipient of "the 2020 World Artificial Intelligence Conference Youth Outstanding Paper" (WAICYOP) award. His research interests include: graph learning, spatio-temporal data mining, reinforcement learning and data-driven traffic simulation.

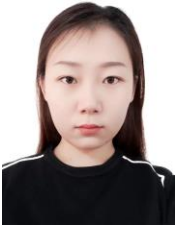

**Dongyu Zuo** is a student member of CCF and is currently pursuing a Master's degree in Software Engineering at Liaoning Technical University in Huludao, Liaoning Province, China.

Her research interests include: reinforcement learning, intelligent vehicle decision-making, and traffic simulation.